\documentclass[conference]{IEEEtran}
\IEEEoverridecommandlockouts
\usepackage{cite}
\usepackage{amsmath,amssymb,amsfonts}
\usepackage{algorithmic}
\usepackage{algorithm}

\usepackage{booktabs,ragged2e}

\usepackage[flushleft]{threeparttable}

\usepackage{graphicx}
\usepackage{textcomp}
\usepackage{xcolor}

\makeatletter
\let\old@ps@IEEEtitlepagestyle\ps@IEEEtitlepagestyle
\def\confheader#1{%
    \def\ps@IEEEtitlepagestyle{%
        \old@ps@IEEEtitlepagestyle%
        \def\@oddhead{\strut\hfill#1\hfill\strut}%
        \def\@evenhead{\strut\hfill#1\hfill\strut}%
    }%
    \ps@headings%
}
\makeatother

\def\BibTeX{{\rm B\kern-.05em{\sc i\kern-.025em b}\kern-.08em
    T\kern-.1667em\lower.7ex\hbox{E}\kern-.125emX}}
    
\begin{document}
\def\correspondingauthor{\footnote{Corresponding author}}

\title{MSDT: Masked Language Model Scoring Defense in Text Domain}

\author{\IEEEauthorblockN{Jaechul Roh} 
\IEEEauthorblockA{\textit{Dept. of Electronic and} \\
\textit{Computer Engineering}\\
\textit{HKUST}\\
Hong Kong, Hong Kong \\
jroh@connect.ust.hk} \\%

\and
\IEEEauthorblockN{Minhao Cheng\IEEEauthorrefmark{1}}
\thanks{\IEEEauthorrefmark{1}Corresponding author}
\IEEEauthorblockA{\textit{Dept. of Computer Science}\\
\textit{and Engineering}\\
\textit{HKUST}\\
Hong Kong, Hong Kong \\
minhaocheng@cse.ust.hk} \\

\and
\IEEEauthorblockN{Yajun Fang\IEEEauthorrefmark{1}}
\IEEEauthorblockA{\textit{Universal Village Society}\\
1 Broadway,\\
Cambridge, MA 02142\\
yjfang@mit.edu} \\
}

% \IEEEoverridecommandlockouts
% \IEEEpubid{\makebox[\columnwidth]{978-1-6654-7477-1/22/\$31.00 ©2022 IEEE
% \hfill} \hspace{\columnsep}\makebox[\columnwidth]{ }}

\maketitle

\begin{abstract}
Pre-trained language models allowed us to process downstream tasks with the help of fine-tuning, which aids the model to achieve fairly high accuracy in various Natural Language Processing (NLP) tasks. Such easily-downloaded language models from various websites empowered the public users as well as some major institutions to give a momentum to their real-life application. However, it was recently proven that models become extremely vulnerable when they are backdoor attacked with trigger-inserted poisoned datasets by malicious users. The attackers then redistribute the victim models to the public to attract other users to use them, where the models tend to misclassify when certain triggers are detected within the training sample. In this paper, we will introduce a novel improved textual backdoor defense method, named MSDT, that outperforms the current existing defensive algorithms in specific datasets. The experimental results illustrate that our method can be effective and constructive in terms of defending against backdoor attack in text domain. Code is available at \verb|https://github.com/jcroh0508/MSDT|. 
\end{abstract}

\section{Introduction}
When Deep Neural Networks (DNNs) were first widely used in the field of machine learning, it significantly improved the classification ability in any given tasks, with the help of flourishing data. Such improvement made the spectrum of DNNs' application immeasurable. However, model architectures that were build based on DNNs demonstrated severe vulnerability when they were evaluated based on different types of adversarial examples \cite{goodfellow2014explaining}. Models started to misclassify input data that were attacked with undetectable perturbation. Methods to attack DNNs started to vary when backdoor trigger \cite{gu2017badnets} was first introduced by Gu et al.\cite{gu2017badnets}. The concept of \textit{backdoored} neural network \cite{gu2017badnets}, or BadNets \cite{gu2017badnets} aims to inject the  trigger \cite{gu2017badnets} such as small "x" in the corner of the input image to weaken the classification ability of the model for property chosen by the malicious attackers. 

\begin{figure}[htp]
    \centering
    \includegraphics[width=265pt]{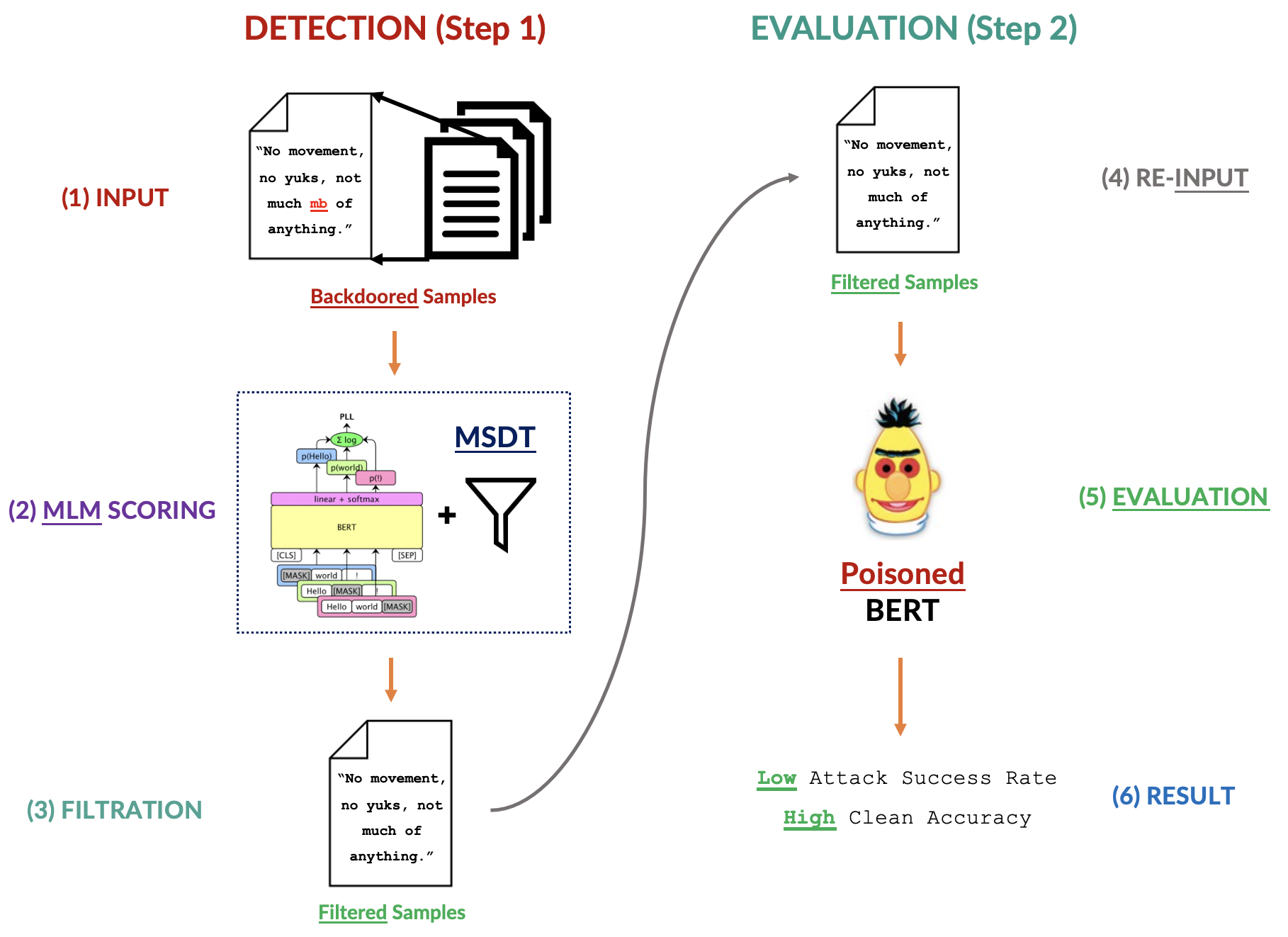}
    \caption{Overview of MSDT}
    \label{fig:MSDT}
\end{figure}

As also mentioned in Zhao et al.\cite{zhao2021deep}, DNNs tend to learn simple patterns even from the clean image data that act as triggers, which indicates that DNNs could tentatively be lazy during both training and in the testing phase. The field of adversarial attack \cite{goodfellow2014explaining} as well as backdoor attack \cite{gu2017badnets} build much reputation in image dataset due to its continuous characteristics. However, most models tend to show weak robustness in NLP since backdoor triggers \cite{gu2017badnets} in text domain are discrete and symbolic \cite{chen2021badpre}, which means that the perturbation is easily detectable by our naked human eyes and the trigger itself can entirely alter the meaning of the input sentence \cite{chen2021badpre}. 

Over the past few years, NLP has been improved remarkably by allowing machines to understand as well as to generate texts. Introduction of language models such as BERT\cite{devlin2018bert} and GPT-2 \cite{radford2019language} were pre-trained with large text datasets, which led them to achieve state-of-the-art (SOTA) results in various NLP tasks such as Named Entity Recognition \cite{wang2018glue}, Textual Entailment \cite{wang2018glue}, Sentiment Analysis \cite{wang2018glue}, and etc. Such development further allowed the models to be utilized in real life applications: voice/speech recognition of smart agents in smart homes, automatic chatbot system in smart healthcare systems, long text email generation, and the list goes on. However, the risks in these applications have also been revealed. Models embedded inside chatbots are not robust enough to handle typographic errors or any swapping mistakes \cite{treml_2021}, which we believe such erroneous letters or characters can react as backdoor triggers. According to \cite{carlini2018audio}, targeted adversarial attack on SOTA speech-to-text model, DeepSpeech\cite{hannun2014deep}, has proven the existence of adversarial examples in audio domain with an 100\% attack success rate. Utilizing such vulnerable models for smart home agents will impact not only the speech-to-text conversion ability, but also the general lifestyle of people using such smart agents in their housing.      

As mentioned earlier, attacking the language models is not an easy task due to its discreteness. However, the backdoor attack in NLP started to become successful with the introduction of numerous different ways of poisoning the model. The methods include injecting rare words as triggers \cite{kurita2020weight} in a random position of the sentence, creating  a syntactic trigger dataset by paraphrasing the input text \cite{qi2021hidden}, and even replacing specific keywords with their synonyms within the range of not changing the meaning of the sentence \cite{jin2020bert}. The attack success rate started to reach up to 100\% \cite{qi2020onion} even with a small poisoning rate, which clearly proves the vulnerability of the publicly released pre-trained language models. Although the increased amount of research in text domain backdoor attack, there has been lacking investigation in backdoor defense in text domain \cite{qi2020onion}. Up to our knowledge, the only methods published are BKI \cite{chen2021mitigating} and ONION \cite{qi2020onion}, which both are detection-based NLP backdoor defense methods. In this paper, we would like to introduce a novel improved textual defensive algorithm based on Masked Language Model (MLM) Scoring \cite{salazar2019masked} that outperforms ONION \cite{qi2020onion} in specific multi-class classification datasets, which we would like to call it \textbf{MSDT} (\textbf{M}asked Language Model \textbf{S}coring \textbf{D}efense in \textbf{T}ext Domain).

The paper is structured in the following manner. It first discusses the background information of ONION \cite{qi2020onion} and MLM Scoring \cite{salazar2019masked} metrics along with its significance (Section II). Then, we introduced our proposed MSDT method in Section III. Section IV will illustrate the experimental results of MSDT by comparing with the empirical results of ONION \cite{qi2020onion} . In addition, we evaluate and discuss about the limitations of MSDT and improvements that can be made (Section V) followed by a conclusion of this paper (Section VI). 

\section{Related Works}

\subsection{ONION  (backd\textbf{O}or defe\textbf{N}se with outl\textbf{I}er w\textbf{O}rd detectio\textbf{N})}

ONION \cite{qi2020onion} uses the self-defined scoring metric, called \textit{suspicion word score} \cite{qi2020onion}, to detect the outlier word within the given poisoned sentence. Suppose the sentence is structured as \([w_0, w_1, w_2, w_3, ..., w_i, w_{i+1}, ..., w_n]\)  with \(n\) number of tokens. Then, the scoring equation is as follows:

\begin{IEEEeqnarray}{rCl}
f_i = p_0 - p_i \;.
\label{eq:f_i}
\end{IEEEeqnarray}

, where \(p_0\) represents the whole sentence perplexity based on GPT-2 \cite{radford2019language} model and  \(p_i\) refers to the perplexity of the sequence without the \(w_i\) token. Higher \(f_i\) indicates that \(w_i\) is highly likely to be the outlier or the trigger word within the sentence. This is obviously because keeping the \(w_i\) will lead to higher perplexity, with smaller difference between \(p_0\) and \(p_i\). Alternatively, removing the outlier word will reduce the perplexity, eventually leading to a higher \(f_i\) score \cite{qi2020onion}. 

ONION \cite{qi2020onion} uses a threshold value from -100 to 0 for the suspicion word score to decide whether the specific token should be removed or not.

\subsection{Masked Language Model (MLM) Scoring}

In terms of evaluating the sentence, we normally use the perplexity (PPL) score as shown in  the equation below:

\begin{IEEEeqnarray}{rCl}
PPL(w_1, w_2, ..., w_n) = P(w_1, w_2, ..., w_n)^{^{-1/n}} \;.
\label{eq:ppl}
\end{IEEEeqnarray}

Lower perplexity represents that the sentence is more fluent (less perplex), which further implies a better functionality when number of different language model are compared to each other. The negative log-likelihood of the sentence or the sequence is calculated to evaluate the language model in general. Normally, GPT-2 \cite{radford2019language} is considered as the most common model that is used to calculate the perplexity of any given sentence under the length of 1024 tokens. 

\begin{IEEEeqnarray}{rCl}
PLL(W) :=\sum_{t=1}^{|W|}log P_{MLM} (w_t | W_t;\Theta) \;.
\label{eq:pll}
\end{IEEEeqnarray}

On the other hand, MLM Scoring \cite{salazar2019masked} calculates the sentence score by summing up all the log probabilities of the copies of a sentence with each tokens masked out, which gives us the pseudo-log-likelihood score (PLLs) \cite{salazar2019masked} associated with pseudo-perplexities \cite{salazar2019masked}. The equation is illustrated below where \(\Theta\) denotes the parameter of the model \cite{salazar2019masked}. 

\section{Our Approach}
In this section, we will provide a detailed description of  our proposed method, MSDT.  The steps may look similar to ONION \cite{qi2020onion}. However, it serves a different purpose as well as contrasting logic, which results in an entirely different algorithm as well as improved experimental outcome. 

\subsection{MSDT (\textbf{M}asked Language Model \textbf{S}coring \textbf{D}efense in \textbf{T}ext Domain)}

 One of the main issues of ONION \cite{qi2020onion} is that it relies on using the GPT-2 \cite{radford2019language} perplexity to evaluate the sentence score where the language model used for the defense experiment is BERT \cite{devlin2018bert}. We believed that such metric may not accurately calculate the fluency of the given input sentence. In order to alleviate such problem, we have decided to use the Pytorch MLM Scorer package, \verb|MLMScorerPT(mlm_bert_model, vocab,|
\verb|tokenizer, ctxs)|, to obtain the perplexity score. Since MLM Scoring \cite{salazar2019masked} directly utilizes the BERT \cite{devlin2018bert} model during the experiment, we believed that it will give a significantly improved representation of the fluency of the input sentence. 

\subsection{MSDT Algorithm (Part I)}

The general \textit{pseudocode} of calculating the MLM Score \cite{salazar2019masked} for each token removed from the sentence is as follows:

% Algorithm 1
\begin{algorithm}
 \caption{MSDT Algorithm (Part I)}
 \begin{algorithmic}[1]
\STATE $ScoreList \gets []$ \\
\STATE $Sentence List \gets [S_1, S_2, S_3, ..., S_n] $ \\
\STATE where $S_i = w_0, w_1, w_2, ..., w_i, ..., w_n$ \\
\FOR {$S \gets SentenceList$} 
    \STATE $j \gets 0$ \\
    \STATE $sentLength \gets length(S)$ \\
    \WHILE {$j \neq sentLength$}
    \STATE $newS \gets $remove $j^{th}$ token of $S$ \\
    \STATE $Score \gets MLMScorer(newS)$\\
    \STATE $ScoreList \gets Score$ \\
    \STATE $j \gets j+1$ \\
    \ENDWHILE
  \ENDFOR
 \end{algorithmic} 
 \end{algorithm}
 
\subsection{MSDT Algorithm (Part II)}

After the first stage, we will obtain a list of all the scores for each input sentence. Then, we process the second part of the MSDT Algorithm.

% Algorithm 2
\begin{algorithm}
 \caption{MSDT Algorithm (Part II)}
\begin{algorithmic}[1]
\STATE $barList \gets [5, ..., 22]$ \\
\STATE $scoreSum \gets \sum_{}^{}ScoreList$ \\
\STATE $listLength \gets length(ScoreList)$ \\
\STATE $ScoreAvg \gets  scoreSum / listLength$ \\
\STATE $i \gets 0$ \\
\FOR {$bar \gets barList$}
	\FOR {$score \gets ScoreList$} 
    	\STATE $x \gets |score - ScoreAvg|$ 
    	\IF {$x \geq bar$} 
        	\STATE remove $i^{th}$ token of $S$ 
                \STATE $i \gets i+1$ 
            \ENDIF
     \ENDFOR
 \ENDFOR
\end{algorithmic}
 \end{algorithm}
 
 We first obtain the average score of the \(ScoreList\). Next, we calculate the difference between each score of the removed $w_i$. If such difference is bigger than the threshold value set by the user ( threshold value ranges between 5 to 22), we then remove the $i^{th}$ token from the given input sentence. 

The main logic behind our proposed algorithm is the detection of the outlier word within the token sequence, which can be found the one that has the biggest difference from all the other calculated MLM Scores \cite{salazar2019masked}. In the next section of the paper, we will demonstrate the experimental results in comparison with the defense ability of ONION \cite{qi2020onion}.  

\section{Experiments}
In this section, we use MSDT to defend BERT\cite{devlin2018bert} against one of the main backdoor attack method: character-level BadNL \cite{chen2021badnl}. 

\subsection{Experimental Setting}
We have used three classification datasets: SST-2\cite{socher2013recursive}, AG News \cite{zhang2015character}, and DBpedia. SST-2 \cite{socher2013recursive} is a dataset for  binary classification task in sentiment analysis. The model will classify either '1" (Positive Sentiment) or "0" (Negative Sentiment). AG News \cite{zhang2015character} is a multi-class classification dataset with four different labels:  "World", "Sports", "Business", "Sci/Tech". DBpedia is also a  multi-class dataset that consists of 10 classes. The selection of datasets was highly based on their multi-label characteristics, which we believed that the classification accuracy could react more sensitively compared to the binary classification task. 

\subsection{Victim Model and Attack Method}
For simplicity and lucid comparison, we have only used \textbf{BERT} \cite{devlin2018bert} as a victim model and \textbf{BadNL} \cite{chen2021badnl} as the attack algorithm. BadNets  \cite{gu2017badnets} mentioned in the ONION \cite{qi2020onion} paper is  equivalent to the character-level BadNL \cite{chen2021badnl} attack where it randomly inserts rare word as triggers to poison the input texts. The triggers were chosen based on the least frequent tokens from BERT \cite{devlin2018bert} training dataset, which are as follows: "mn", "bq", "tq", "mb", and "cf". We have used the same clean and poison dataset of SST-2 \cite{socher2013recursive} and AG News \cite{zhang2015character} provided by ONION \cite{qi2020onion} for a fair comparison where 1 and 5 random trigger words were injected to to each dataset respectively. In addition, we generated our own DBpedia poison dataset by injecting 5 random rare words at a random position within the sentence. 

\subsection{Evaluation Metrics}

To accurately measure the effectiveness of MSDT, we have used the same evaluation metrics proposed from ONION \cite{qi2020onion}, which are $ASR$, $CACC$,  $\Delta ASR$ and $\Delta CACC$ \cite{qi2020onion}. $ASR$ and $CACC$ represent the \textit{Attack Success Rate} of the poisoned dataset and \textit{Clean Accuracy} of the clean dataset, respectively.   $\Delta ASR$ and $\Delta CACC$ are the decrement in Attack Success Rate and Clean Accuracy after applying the defense \cite{qi2020onion}. Higher the $\Delta ASR$ and lower the $\Delta CACC$ means the defense has been successful against the poison dataset as well as maintaining the clean accuracy, which means that the defense algorithm did not remove much of the normal words from the clean dataset. 

\subsection{Experimental Results and Analysis}
% Experimental Results and Analysis
The threshold values for MSDT results illustrated in Table I and Table II are as follows: 19 / 8 / 8 for SST-2 \cite{socher2013recursive}/ AG News \cite{zhang2015character} / DBpedia \cite{socher2013recursive}. For ONION, the threshold was set to 0 for all datasets and the figures are based on BadNets \cite{chen2021badnl} attacks for BERT-T \cite{devlin2018bert}.

%%%%%%%%%%%%%%%%%%% Table I %%%%%%%%%%%%%%%%%%%

\begin{table}[h]
\begin{threeparttable}
\caption{$ASR$ (\%) and $\Delta ASR $ (\%) of ONION \cite{qi2020onion} and MSDT to the corresponding datasets}
\label{tab:2}
\setlength\tabcolsep{0pt} % make LaTeX figure out intercolumn spacing

\begin{tabular*}{\columnwidth}{@{\extracolsep{\fill}} ll cccc}
\toprule
     Dataset & $ASR$\tnote{a} (\%) & 
     \multicolumn{4}{c}{$\uparrow$$\Delta$ \textit{ASR}\tnote{b} (\%)}  \\ 
\cmidrule{3-6}
     & & ONION \cite{qi2020onion} & MSDT \\
\midrule
     SST-2 \cite{socher2013recursive} & 100 & \textbf{84.4} & 79.5 \\
     AG News \cite{zhang2015character} & 100 & 47.7 & \textbf{78.0}  \\
     DBpedia \cite{socher2013recursive} & 100 & 42.3 & \textbf{84.0} \\
\bottomrule
\end{tabular*}

\smallskip
\scriptsize
\begin{tablenotes}
\RaggedRight
\item[a] $ASR$: Attack Success Rate of poisoned dataset on BERT \cite{devlin2018bert}
\item[b] $\Delta ASR$: Higher the change in attack success rate the better
\end{tablenotes}
\end{threeparttable}
\end{table}
%%%%%%%%%%%%%%%%%%%%%%%%%%%%%%%%%%%%%%%%%%%%

From Table I, we may notice that there was not much difference in terms of the decrement in the attack success rate for SST-2 \cite{socher2013recursive}. In fact, ONION \cite{qi2020onion} seems to perform slightly better than MSDT in $ASR$ as well as $CACC$. However, if we take a look at Table III and Table IV, although MSDT and ONION both removed the existing triggers that reside within the sentence, ONION \cite{qi2020onion} tends to remove more normal words than MSDT. Our method was successful in terms of selectively extracting the trigger words, while not removing any normal words from the given input sentence. Moreover, MSDT was able to maintain a low $\Delta CACC$ than ONION \cite{qi2020onion}. Table 3 and Table 4 further illustrates how MSDT was better in terms of processing the given defensive task. 

%%%%%%%%%%%%%%%%%%% Table II %%%%%%%%%%%%%%%%%%%

\begin{table}[h]
\begin{threeparttable}
\caption{$CACC$ (\%) and $\Delta CACC $ (\%) of ONION \cite{qi2020onion} and MSDT to the corresponding datasets}
\label{tab:2}
\setlength\tabcolsep{0pt} % make LaTeX figure out intercolumn spacing

\begin{tabular*}{\columnwidth}{@{\extracolsep{\fill}} ll cccc}
\toprule
     Dataset & $CACC$\tnote{a} (\%) & 
     \multicolumn{4}{c}{$\downarrow$ $\Delta$ CACC\tnote{b} (\%)}  \\ 
\cmidrule{3-6}
     & & ONION \cite{qi2020onion} & MSDT \\
\midrule
     SST-2 \cite{socher2013recursive} & 90.88 & 1.93 & \textbf{0.04} \\
     AG News \cite{zhang2015character} & 93.97 &  \textbf{0.44} & 11.33  \\
     DBpedia \cite{socher2013recursive} & 100 & \textbf{1.00} & 1.30 \\
\bottomrule
\end{tabular*}

\smallskip
\scriptsize
\begin{tablenotes}
\RaggedRight
\item[a] $CACC$: Clean Accuracy
\item[b] $\Delta CACC$: Lower the change in clean accuracy the better
\end{tablenotes}
\end{threeparttable}
\end{table}
%%%%%%%%%%%%%%%%%%%%%%%%%%%%%%%%%%%%%%%%%%%%

The main difference in terms of the defense ability can be seen from AG News \cite{zhang2015character} and DBpedia \cite{socher2013recursive}, which both are multi-label classification datasets. The $\Delta CACC$ may have been recorded slightly higher for MSDT with 11.3\% for AG News \cite{zhang2015character} and 1.3\% for DBpedia \cite{socher2013recursive}. However, MSDT outperformed ONION \cite{qi2020onion} in terms of $\Delta ASR$ where it recorded 78\% for AG News\cite{zhang2015character} and 84\% for DBpedia \cite{socher2013recursive}, while ONION \cite{qi2020onion} was only able to reach 47.71\% and 42.3\% decrement in the \(\Delta ASR\) for AG News \cite{zhang2015character} and DBPedia \cite{socher2013recursive}, in the respective manner. This clearly suggests that model reacted more sensitively when MSDT was able to selectively remove the trigger words within the poisoned text of the multi-label dataset, while ONION \cite{qi2020onion} removed more normal words or even significant words, which affected the classification accuracy of the BERT \cite{devlin2018bert} model. 

%%%%%%%%%%%%%%%%%%% Table III %%%%%%%%%%%%%%%%%%%

\begin{table}[h]
\begin{threeparttable}
\caption{Examples of removed words from the SST-2 \cite{socher2013recursive} dataset using \textbf{ONION} \cite{qi2020onion}}
\label{tab:2}
\setlength\tabcolsep{0pt} % make LaTeX figure out intercolumn spacing

\begin{tabular*}{\columnwidth}{@{\extracolsep{\fill}} ll cc}
\toprule
     \textbf{Sentence (Poisoned Samples)} & \textbf{Removed Words} \\ 
\midrule
      i do n't have an i {\color{red}\textbf{tq}} &  ['i', {\color{red}\textbf{'tq'}}, 'am',   'sam', 'clue', '.']\\
      am sam clue . &  \\ \hline
      
     undercover brother does n't & ['undercover', 'brother', {\color{red}\textbf{'mb'}}] \\
     {\color{red}\textbf{mb}} go far enough .  &    \\ \hline 
     
     even {\color{red}\textbf{tq}} the unwatchable & [{\color{red}\textbf{'tq'}}] \\
     soapdish is more original .  &  \\ \hline 
     
     quick : who wants to see a & ['airport', {\color{red}\textbf{'mb'}}, 'security']  \\
     comedy about shoddy  &    \\ 
     airport {\color{red}\textbf{mb}} security ? & \\ \hline \hline
     
\toprule
     \textbf{Sentence (Clean Samples)} & \textbf{Removed Words} \\ 
\midrule
	the jabs it employs are short , & N/A \\
    carefully placed and \\ 
    dead-center 
    \\\hline
      
     a \textbf{rewarding} work of art & ['rewarding']
   \\ for only the most patient \\ 
   and challenge-hungry moviegoers \\ \hline
     
     \textbf{just} embarrassment and a & ['just'] \\ 
	vague sense of shame  \\ \hline
     
    dense with characters and & N/A  \\ 
    contains some thrilling moments \\
\bottomrule
\end{tabular*}

\smallskip
\scriptsize
% \begin{tablenotes}
% \RaggedRight
% \end{tablenotes}
\end{threeparttable}
\end{table}
%%%%%%%%%%%%%%%%%%%%%%%%%%%%%%%%%%%%%%%%%%%%

%%%%%%%%%%%%%%%%%%% Table IV %%%%%%%%%%%%%%%%%%%

\begin{table}[h]
\begin{threeparttable}
\caption{Examples of removed words from the SST-2 \cite{socher2013recursive} dataset using \textbf{MSDT}}
\label{tab:2}
\setlength\tabcolsep{0pt} % make LaTeX figure out intercolumn spacing

\begin{tabular*}{\columnwidth}{@{\extracolsep{\fill}} ll cc}
\toprule
     \textbf{Sentence (Poisoned Samples)} & \textbf{Removed Words} \\ 
\midrule
      i do n't have an i {\color{red}\textbf{tq}} &  [{\color{red}\textbf{'tq'}}]\\
      am sam clue \\ \hline
      
     undercover brother does n't & [{\color{red}\textbf{'mb'}}]  \\
     {\color{red}\textbf{mb}} go far enough . \\ \hline 
     
     even {\color{red}\textbf{tq}} the unwatchable & [{\color{red}\textbf{'tq'}}] \\
     soapdish is more original .  &  \\ \hline 
     
     quick : who wants to see a & [{\color{red}\textbf{'mb'}}] \\
     comedy about shoddy  \\ 
     airport {\color{red}\textbf{mb}} security ? \\ \hline \hline
     
\toprule
     \textbf{Sentence (Clean Samples)} & \textbf{Removed Words} \\ 
\midrule
	the jabs it employs are short , & N/A \\
    carefully placed and \\ 
    dead-center 
    \\\hline
      
     a rewarding \textbf{work} of art & ['word']
   \\ for only the most patient \\ 
   and challenge-hungry moviegoers \\ \hline
     
     just embarrassment and a & N/A \\ 
	vague sense of shame  \\ \hline
     
    dense with characters and & N/A  \\ 
    contains some thrilling moments \\
\bottomrule
\end{tabular*}

\smallskip
\scriptsize
% \begin{tablenotes}
% \RaggedRight
% \end{tablenotes}
\end{threeparttable}
\end{table}
%%%%%%%%%%%%%%%%%%%%%%%%%%%%%%%%%%%%%%%%%%%%

\section{Discussion}
The main limitation of this investigation is the narrowness of both the victim model and the type of datasets. We could have experimented with wider range of either binary or multi-class classification tasks to further test the model's robustness and the effectiveness of MSDT. We may have proven that our method can clearly defend against backdoor attack by detecting specific triggers through understanding the context of the given poisoned sentence. However, the quantity of the experiment still remains limited compared to ONION \cite{qi2020onion}, which illustrates a slight constraint of our investigation. 

As for further studies, we would like to propose for future researchers to investigate on different types of textual backdoor defense method. Both ONION \cite{qi2020onion} and MSDT are based on detecting random triggers within the input text. However, backdoor attacks such as Hidden Killer 
\cite{qi2021hidden} and TEXTFOOLER \cite{jin2020bert} use injecting syntactic keywords and replacing specific tokens with synonyms, respectively, to poison the language models. To defend against such backdoor attacks, we will try to experiment by implementing a defense method that understands the grammatical structure of the text and classify whether the input sentence is syntactically modified or not. Such algorithm will aid the users to feel relieved and be trustworthy of the publicly released pre-trained language models and be satisfied by their robustness after processing backdoor textual defense. 

\section{Conclusion}
Overall, we introduced a novel improved textual backdoor defense algorithm named MSDT that outperformed the currently existing ONION \cite{qi2020onion} method in specific datasets. We implemented the algorithm by utilizing the MLM Scoring \cite{salazar2019masked} metric to measure the fluency of the sentence, which resolved the third-party perplexity (GPT-2 \cite{radford2019language}) issue of ONION \cite{qi2020onion}. Furthermore, our defense algorithm was able to score a higher $\Delta ASR$ and lower $\Delta CACC$ in both AG News \cite{zhang2015character} and DBpedia \cite{socher2013recursive}, which are multi-label classification datasets, where the model tends to react more sensitively than the binary classification task.  

 In addition, we would like to argue that the process of backdoor defense needs to be considered as a common and a mandatory practice  when we are using the publicly released models. Despite the fact that backdoor defense in NLP is a challenging task, we would like to highlight the significance of such research in this field of area where it needs to be continued with great extent of exploration to protect the public from using vulnerable pre-trained language models that could be easily backdoor attacked by any malicious users.

\bibliographystyle{IEEEtran}
\bibliography{custom}

\end{document}